\documentclass[10pt]{article}

\usepackage{eaclap}
\def\wordcount#1{\gdef\wc{#1}}
\gdef\lastpage{1}

\usepackage{psfig}
\usepackage{piclatex}
\usepackage{rotate,tools}

\author{Thorsten Brants\\
        Universit\"at des Saarlandes, Computerlinguistik\\
        D-66041 Saarbr\"ucken, Germany\\
        {\tt thorsten@coli.uni-sb.de}\\[1ex]
        In {\em Proceedings of 9th Conference of the {E}uropean Chapter
        of the}\\ {\em {A}ssociation for {C}omputational {L}inguistics
        {EACL}-99}. Bergen, Norway, 1999}

\title{Cascaded Markov Models}

\wordcount{2800}

\def\stern{\ifmmode^{\ast}\else$^{\ast}$\fi}
\def\argmax{\mathop{\rm argmax}}
\def\N{\mbox{I\kern-.2emN}}

\begin{document}

\maketitle

\begin{abstract}
  This paper presents a new approach to partial parsing
  of context-free structures. The approach is based
  on Markov Models. Each layer of the resulting structure is
  represented by its own Markov Model, and output of a lower layer is
  passed as input to the next higher layer. An empirical evaluation
  of the method yields very good results for {\sf NP}/{\sf PP} chunking
  of German newspaper texts.
\end{abstract}



\section{Introduction}

Partial parsing, often referred to as {\em chunking}, is used as a
pre-processing step before deep analysis or as shallow processing for
applications like information retrieval, messsage extraction and text
summarization.  Chunking concentrates on constructs that can
be recognized with a high degree of certainty. For several
applications, this type of information with high accuracy is more
valuable than deep analysis with lower accuracy.

We will present a new approach to partial parsing that uses Markov Models.
The presented models are extensions of the part-of-speech tagging
technique and are capable of emitting structure.
They utilize context-free grammar rules and add
left-to-right transitional context information. This type of model is
used to facilitate the syntactic annotation of the NEGRA corpus
of German newspaper texts \cite{Skut:ea:97a}.

Part-of-speech tagging is the assignment of syntactic categories (tags)
to words that occur in the processed text. Among others, this task is
efficiently solved with Markov Models. States of a Markov Model
represent syntactic categories (or tuples of syntactic categories), and
outputs represent words and punctuation \cite[and
others]{Church:88,DeRose:88}. This technique of statistical
part-of-speech tagging operates very successfully, and usually accuracy
rates between 96 and 97\% are reported for new, unseen text.

\newcite{Brants:ea:97} showed that the technique of statistical
tagging can be shifted to the next level of syntactic processing and
is capable of assigning grammatical functions. These are
functions like {\em subject}, {\em direct object}, {\em head}, etc.
They mark the function of a child node within its parent phrase.

\begin{figure*}
\hrule
\smallskip

\dimen0=\textwidth\advance\dimen0 by 5mm%
\psfig{file=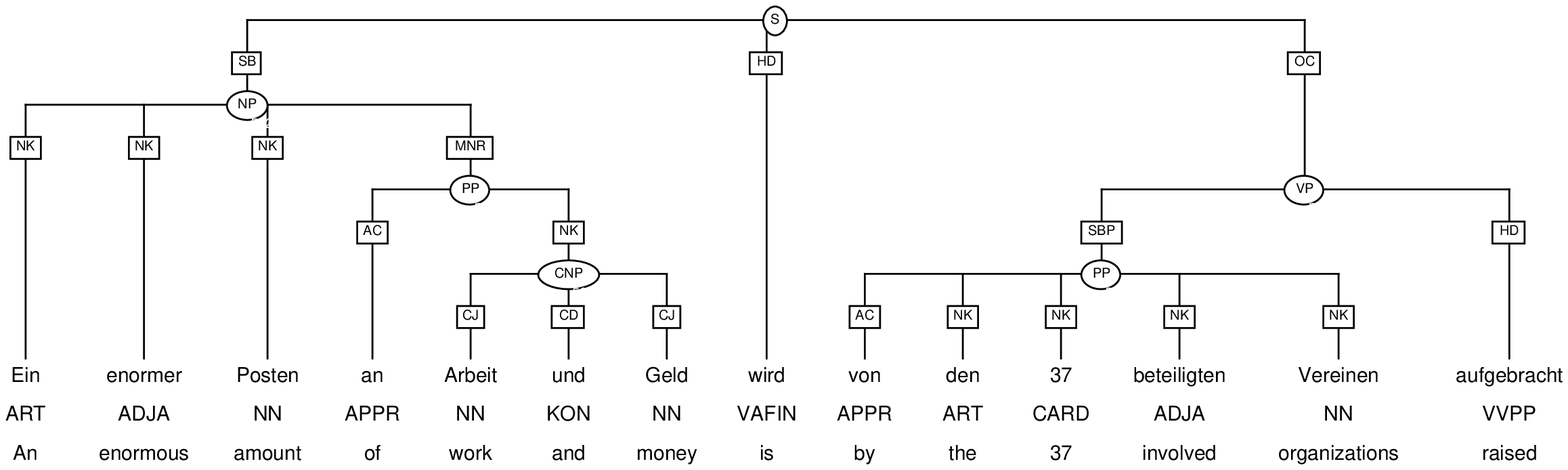,width=\dimen0}
\vspace*{-2ex}
\centerline{\em `A large amount of money and work was raised by the
  involved organizations'}
\smallskip
\hrule

\caption{Example sentence and annotation. The structure consists of
  terminal nodes (words and their parts-of-speech), non-terminal nodes
  (phrases) and edges (labeled with grammatical functions).}
\label{fig:examplesent}
\end{figure*}

Figure \ref{fig:examplesent} shows an example sentence and its
structure. The terminal sequence is com\-ple\-men\-ted by tags
(Stuttgart-T{\"u}bingen-Tagset, Thielen and Schiller, 1995).
\nocite{Thielen:Schiller:94} Non-terminal nodes are labeled with phrase
categories, edges are labeled with grammatical functions (NEGRA tagset).


In this paper, we will show that Markov Models are not restricted to
the labeling task (i.e., the assignment of part-of-speech labels,
phrase labels, or labels for grammatical functions), but are also
capable of generating structural elements. We will use cascades of
Markov Models. Starting with the part-of-speech layer, each layer of
the resulting structure is represented by its own Markov Model. A
lower layer passes its output as input to the next higher layer. The
output of a layer can be ambiguous and it is complemented by a
probability distribution for the alternatives.

This type of parsing is inspired by finite state cascades which are
presented by several authors.

CASS \cite{Abney:91,Abney:96} is a partial parser that recognizes 
non-recursive basic phrases (chunks) with finite state transducers. Each
transducer emits a single best analysis (a longest match) that serves
as input for the transducer at the next higher level. CASS
needs a special grammar for which rules are manually coded. Each layer
creates a particular subset of phrase types.
FASTUS \cite{Appelt:ea:93} is heavily based on pattern matching. Each
pattern is associated with one or more trigger words. It uses a series of
non-deterministic finite-state transducers to build chunks; the output of
one transducer is passed as input to the next transducer.
\cite{Roche:94} uses the fix point of
a finite-state transducer. The transducer is iteratively applied to
its own output until it remains identical to the
input. The method is successfully used for efficient
processing with large grammars. 
\cite{Cardie:Pierce:98} present an approach to chunking based on a
mixture of finite state and context-free techniques. They use {\sf NP}
rules of a pruned treebank grammar. For processing, each point of a
text is matched against the treebank rules and the longest match is
chosen.  Cascades of automata and transducers can also be found in
speech processing, see e.g.\ \cite{Pereira:ea:94,Mohri:97}.

Contrary to finite-state transducers, Cascaded Markov Models exploit
probabilities when processing layers of a syntactic structure.
They do not generate longest matches but most-probable
sequences. Furthermore, a higher layer sees different
alternatives and their probabilities for the same span.
It can choose a lower ranked alternative if
it fits better into the context of the higher layer.  An additional
advantage is that Cascaded Markov Models do not need a
``stratified'' grammar (i.e., each layer encodes a disjoint subset of
phrases).  Instead the
system can be immediately trained on existing treebank data.

The rest of this paper is structured as follows. Section
\ref{sec:coding} addresses the encoding of parsing processes as Markov
Models. Section \ref{sec:cascaded} presents Cascaded Markov
  Models. Section \ref{sec:experiment} reports on the evaluation of
Cascaded Markov Models using treebank data. Finally, section
\ref{sec:conclusion} will give conclusions.


\section{Encoding of Syntactical Information as Markov Models}
\label{sec:coding}

\begin{figure*}
\hrule
\vspace{-5mm}
\centerline{\psfig{file=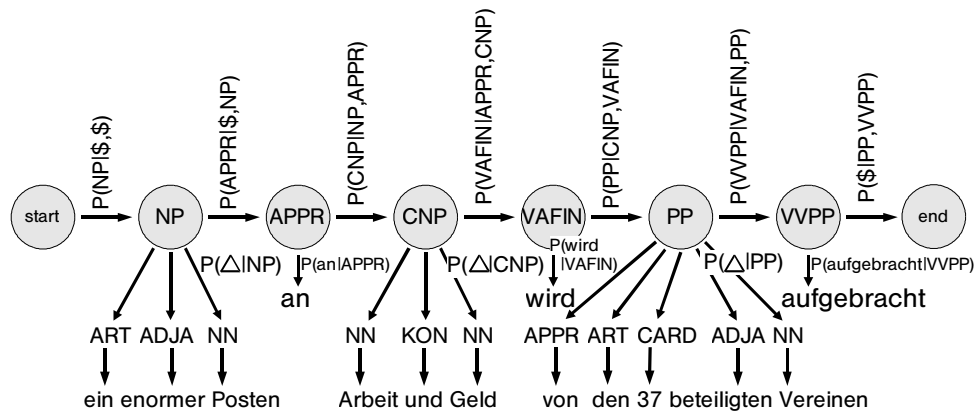,width=13cm,angle=-90}}
\medskip
\hrule
\caption{Part of the Markov Models for layer 1 that is used to process
  the sentence of figure \ref{fig:examplesent}. Contrary to
  part-of-speech tagging, outputs of states may consist of structures with
  probabilities according to a stochastic context-free grammar.}
\label{fig:mm-level1}
\end{figure*}  

When encoding a part-of-speech tagger as a Markov Model, states
represent syntactic categories\footnote{Categories and states directly
  correspond in bigram models. For higher order models, tuples of
  categories are combined to one state.} and outputs represent words.
Contextual probabilities of tags are encoded as transition
probabilities of tags, and lexical probabilities of the Markov Model
are encoded as output probabilities of words in states.

We introduce a modification to this encoding. States additionally may
represent non-terminal categories (phrases).  These new states emit
partial parse trees (cf.~figure \ref{fig:mm-level1}). This can be
seen as collapsing a sequence of terminals into one non-terminal.
Transitions into and out of the new states are performed in the same
way as for words and parts-of-speech.

Transitional probabilities for this new type of Markov Models can be
estimated from annotated data in a way very similar to estimating
probabilities for a part-of-speech tagger. The only difference is that
sequences of terminals may be replaced by one non-terminal.

Lexical probabilities need a new estimation method.
We use probabilities of context-free partial parse
trees. Thus, the lexical probability of the state {\sf NP} in figure
\ref{fig:mm-level1} is determined by 
\begin{eqnarray*}
 & & \hskip-5mm P({\sf NP} \rightarrow {\sf ART\ ADJA\ NN}, \\
 & &      {\sf ART} \rightarrow {\rm ein},
          {\sf ADJA} \rightarrow {\rm enormer},
          {\sf NN} \rightarrow {\rm Posten}) \\
 & = &
  P({\sf NP} \rightarrow {\sf ART\ ADJA\ NN}) \\
 &   & \cdot
  P({\sf ART} \rightarrow {\rm ein}) \cdot
  P({\sf ADJA} \rightarrow {\rm enormer}) \\
 & & \cdot  P({\sf NN} \rightarrow {\rm Posten})
\end{eqnarray*}
Note that the last three probabilities are the same as for the
part-of-speech model.


\section{Cascaded Markov Models}
\label{sec:cascaded}

The basic idea of Cascaded Markov Models is to construct the parse tree
layer by layer, first structures of depth one, then structures of depth
two, and so forth. For each layer, a Markov Model determines the best
set of phrases. These phrases are used as input for the next layer,
which adds one more layer.  Phrase hypotheses at each layer are
generated according to stochastic context-free grammar rules (the
outputs of the Markov Model) and subsequently filtered from left to
right by Markov Models.

\begin{figure*}
\hrule
\medskip
\psfig{file=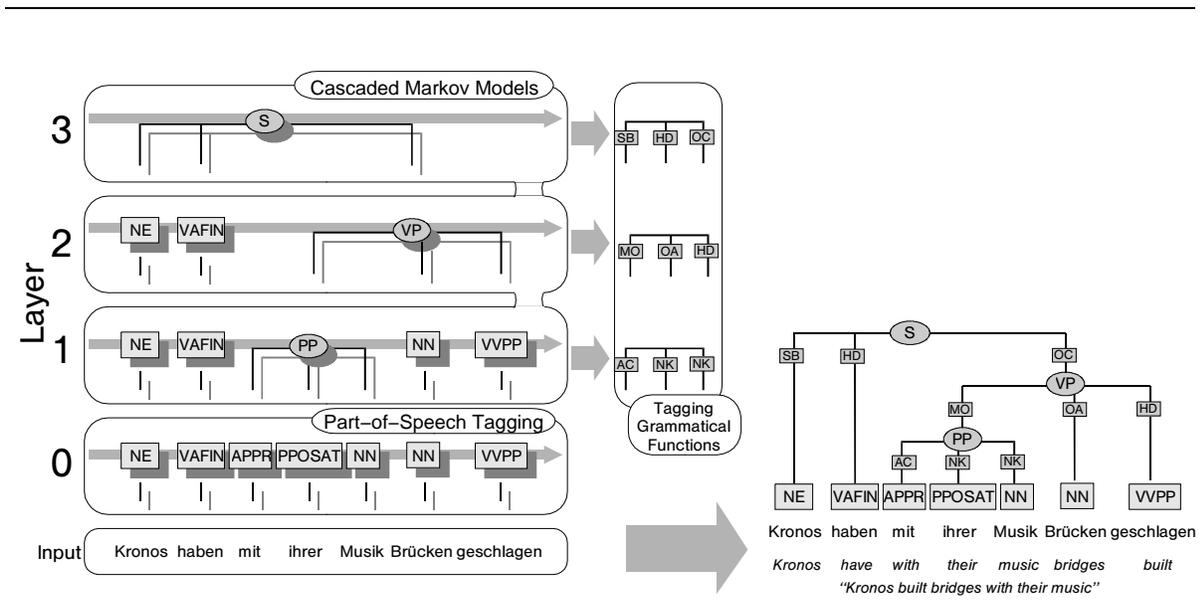,width=\textwidth,angle=-90}
\medskip
\hrule
\caption{The combined, layered processing model. Starting with
  part-of-speech tagging (layer 0), possibly ambiguous output together
  with probabilities is passed to higher layers (only the best
  hypotheses are shown for clarity). At each layer, new
  phrases and grammatical functions are added.}
\label{fig:modelv}
\end{figure*}  

Figure \ref{fig:modelv} gives an overview of the parsing
model. Starting with part-of-speech tagging, new phrases are created at
higher layers and filtered by Markov Models operating from left to right.


\subsection{Tagging Lattices}

The processing example in figure \ref{fig:modelv} only shows the best
hypothesis at each layer. But there are alternative phrase
hypotheses and we need to determine the best one during the parsing
process.

\begin{figure*}
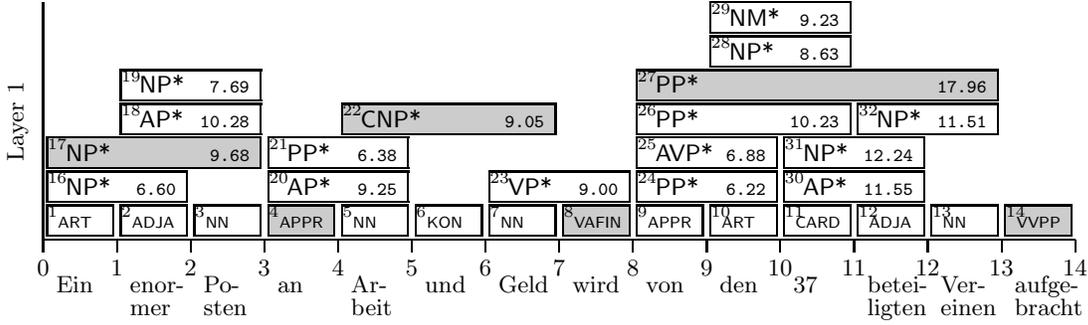

\hrule
\medskip
\begin{center}
\let\termsize=\footnotesize
\let\indexsize=\tiny
\let\posindexsize=\tiny
\let\probsize=\scriptsize
\let\posprobsize=\scriptsize
\let\nodesize=\small
\let\posnodesize=\tiny
\let\childrensize=\scriptsize

\def\xunit{9.8}
\def\yunit{4.5}
\unitlength=1mm

\def\term#1#2{\put{\termsize #2} [lB] <5pt,-7mm> at #1 0}
\def\termrest#1#2{\put{\termsize #2} [lB] <5pt,-10mm> at #1 0}

\def\posnode#1#2#3#4#5#6#7#8{%
  \put{\posindexsize #1} [lt] <.6mm,4.4mm> at #2 #3
  \put{\posnodesize\sf #6} [lb] <2mm,1.5mm> at #2 #3
  \putrectangle <.5mm,.5mm> corners at #2 #3 and #4 #5
}
\def\mposnode#1#2#3#4#5#6#7#8{%
  {\setgray{0.8}{\linethickness=4mm\putbar <.5mm,2.5mm> breadth <0pt> from #2 #3 to #4 #3 }\unsetgray}%
  \put{\posindexsize #1} [lt] <.6mm,4.4mm> at #2 #3
  \put{\posnodesize\sf #6} [lb] <2mm,1.5mm> at #2 #3
  \putrectangle <.5mm,.5mm> corners at #2 #3 and #4 #5
}
\def\node#1#2#3#4#5#6#7#8{%
  \put{\indexsize #1} [lt] <.6mm,4.4mm> at #2 #3
  \put{\nodesize\sf #6} [lb] <3mm,1.5mm> at #2 #3
  \put{\probsize\tt #7} [rb] <-1mm,1.5mm> at #4 #3
  \putrectangle <.5mm,.5mm> corners at #2 #3 and #4 #5
}
\def\mnode#1#2#3#4#5#6#7#8{%
  {\setgray{0.8}{\linethickness=4mm\putbar <.5mm,2.5mm> breadth <0pt> from #2 #3 to #4 #3 }\unsetgray}%
  \put{\indexsize #1} [lt] <.6mm,4.4mm> at #2 #3
  \put{\nodesize\sf #6} [lb] <3mm,1.5mm> at #2 #3
  \put{\probsize\tt #7} [rb] <-1mm,1.5mm> at #4 #3
  \putrectangle <.5mm,.5mm> corners at #2 #3 and #4 #5
}

\mbox{}%
\beginpicture
\setcoordinatesystem units <\xunit mm,\yunit mm>
\setplotarea x from 0 to 14, y from 0 to 7
\small\axis bottom ticks numbered from 0 to 14 by 1 /
\axis left /
\put{\rotate{Layer 1}} [r] <-2mm,0mm> at 0 3.5
\linethickness=.4pt
\term{0}{Ein}
\term{1}{enor-}
\termrest{1}{mer}
\term{2}{Po-}
\termrest{2}{sten}
\term{3}{an}
\term{4}{Ar-}
\termrest{4}{beit}
\term{5}{und}
\term{6}{Geld}
\term{7}{wird}
\term{8}{von}
\term{9}{den}
\term{10}{37}
\term{11}{betei-}
\termrest{11}{ligten}
\term{12}{Ver-}
\termrest{12}{einen}
\term{13}{aufge-}
\termrest{13}{bracht}
\posnode{1}{0}{0}{0.9}{0.9}{ART}{3.40}{}
\node{16}{0}{1}{1.9}{1.9}{NP*}{6.60}{1 2}
\mnode{17}{0}{2}{2.9}{2.9}{NP*}{9.68}{1 2 3}
\posnode{2}{1}{0}{1.9}{0.9}{ADJA}{3.01}{}
\node{18}{1}{3}{2.9}{3.9}{AP*}{10.28}{2 3}
\node{19}{1}{4}{2.9}{4.9}{NP*}{7.69}{2 3}
\posnode{3}{2}{0}{2.9}{0.9}{NN}{5.32}{}
\mposnode{4}{3}{0}{3.9}{0.9}{APPR}{2.90}{}
\node{20}{3}{1}{4.9}{1.9}{AP*}{9.25}{4 5}
\node{21}{3}{2}{4.9}{2.9}{PP*}{6.38}{4 5}
\posnode{5}{4}{0}{4.9}{0.9}{NN}{3.92}{}
\mnode{22}{4}{3}{6.9}{3.9}{CNP*}{9.05}{5 6 7}
\posnode{6}{5}{0}{5.9}{0.9}{KON}{1.80}{}
\posnode{7}{6}{0}{6.9}{0.9}{NN}{4.06}{}
\node{23}{6}{1}{7.9}{1.9}{VP*}{9.00}{7 8}
\mposnode{8}{7}{0}{7.9}{0.9}{VAFIN}{2.51}{}
\posnode{9}{8}{0}{8.9}{0.9}{APPR}{2.49}{}
\node{24}{8}{1}{9.9}{1.9}{PP*}{6.22}{9 10}
\node{25}{8}{2}{9.9}{2.9}{AVP*}{6.88}{9 10}
\node{26}{8}{3}{10.9}{3.9}{PP*}{10.23}{9 10 11}
\mnode{27}{8}{4}{12.9}{4.9}{PP*}{17.96}{9 10 11 12 13}
\posnode{10}{9}{0}{9.9}{0.9}{ART}{2.32}{}
\node{28}{9}{5}{10.9}{5.9}{NP*}{8.63}{10 11}
\node{29}{9}{6}{10.9}{6.9}{NM*}{9.23}{10 11}
\posnode{11}{10}{0}{10.9}{0.9}{CARD}{5.48}{}
\node{30}{10}{1}{11.9}{1.9}{AP*}{11.55}{11 12}
\node{31}{10}{2}{11.9}{2.9}{NP*}{12.24}{11 12}
\posnode{12}{11}{0}{11.9}{0.9}{ADJA}{5.68}{}
\node{32}{11}{3}{12.9}{3.9}{NP*}{11.51}{12 13}
\posnode{13}{12}{0}{12.9}{0.9}{NN}{4.95}{}
\mposnode{14}{13}{0}{13.9}{0.9}{VVPP}{5.25}{}
\endpicture
\end{center}
\hrule
\caption{Phrase hypotheses according to a context-free grammar for
  the first layer. Hypotheses marked with an asterisk (*) are newly
  generated at this layer, the others are passed from the next lower
  layer (layer 0: part-of-speech tagging). The best path in the
  lattice is marked grey.}
\label{fig:cmm-level1}
\end{figure*}

All rules of the generated context-free grammar with right sides that are
compatible with part of the sequence are added to the search space.
Figure \ref{fig:cmm-level1} shows an example for hypotheses at the
first layer when processing the sentence of figure
\ref{fig:examplesent}.  Each bar represents one hypothesis.  The
position of the bar indicates the covered words. It is labeled with the
type of the hypothetical phrase, an index in the upper left corner for
later reference, the negative logarithm of the probability that this
phrase generates the terminal yield (i.e., the smaller the better;
probabilities for part-of-speech tags are omitted for clarity).  This
part is very similar to chart entries of a chart parser.

All phrases that are newly introduced at this layer are marked with an
asterisk (*). They are produced according to context-free rules,  based
on the elements passed from the next lower layer. The layer below
layer 1 is the part-of-speech layer.

The hypotheses form a lattice, with the word boundaries being states
and the phrases being edges. Selecting the best hypothesis means to
find the best path from node 0 to the last node (node 14 in the
example). The best path can be efficiently found with the Viterbi
algorithm \cite{Viterbi:67}, which runs in time linear to the length
of the word sequence.  Having this view of finding the best
hypothesis, processing of a layer is similar to word lattice
processing in speech recognition (cf.\ Samuelsson, 1997).
\nocite{Samuelsson:97}

Two types of probabilities are important when searching for the best
path in a lattice. 
First, these are probabilities of the hypotheses (phrases) generating the
underlying terminal nodes (words). They are calculated according to a 
stochastic context-free grammar and given in figure
\ref{fig:cmm-level1}. The second type are context probabilities, i.e.,
the probability that some type of phrase follows or precedes another.
The two types of probabilities coincide with lexical and contextual
probabilities of a Markov Model, respectively.

According to a trigram model (generated from a corpus), the path
in figure \ref{fig:cmm-level1} that is marked grey is the best path in
the lattice. Its probability is composed of

\[
\begin{array}{l}
   P_{best} = \\
   \ P({\sf NP}|\$,\$) P({\sf NP} \Rightarrow^* \mbox{\em ein enormer Posten}) \\
   \ \cdot P({\sf APPR}|\$,{\sf NP}) P({\sf APPR} \rightarrow an) \\
   \ \cdot P({\sf CNP}|{\sf NP}, {\sf APPR}) P({\sf CNP} \Rightarrow^* \mbox{\em Arbeit und Geld}) \\
   \ \cdot P({\sf VAFIN}|{\sf APPR},{\sf CNP}) P({\sf VAFIN} \rightarrow wird) \\
\end{array}
\]
\[
\begin{array}{l}
   \ \cdot P({\sf PP}|{\sf CNP}, {\sf VAFIN}) \\
   \hspace*{2.6em}P({\sf PP} \Rightarrow^* \mbox{\em von\ den\ 37\ beteiligten\ Vereinen}) \\
   \ \cdot P({\sf VVPP}|{\sf VAFIN},{\sf PP}) P({\sf VVPP} \rightarrow aufgebracht) \\
   \ \cdot P(\$|{\sf PP},{\sf VVPP}). \\
\end{array}
\]

Start and end of the path are indicated by a dollar sign (\$). This
path is very close to the correct structure for layer 1. The {\sf CNP}
and {\sf PP} are correctly recognized. Additionally, the best path
correctly predicts that {\sf APPR}, {\sf VAFIN} and {\sf VVPP} should
not be attached in layer 1. The only error is the {\sf NP} {\em ein
  enormer Posten}. Although this is on its own a perfect {\sf NP}, it
is not complete because the {\sf PP} {\em an Arbeit und Geld} is
missing. {\sf ART}, {\sf ADJA} and {\sf NN} should be left unattached
in this layer in order to be able to create the correct structure at
higher layers.

The presented Markov Models act as {\em filters}. The probability of a
connected structure is determined only based on a stochastic
context-free grammar. The joint probabilities of unconnected partial
structures are determined by additionally using Markov Models. While
building the structure bottom up, parses that are unlikely according to
the Markov Models are pruned.

\subsection{The Method}
\label{sec:cmm-method}

The standard Viterbi algorithm is modified in order to process
Markov Models operating on lattices.  In part-of-speech tagging, each
hypothesis (a tag) spans exactly one word. Now, a hypothesis can span
an arbitrary number of words, and the same span can be covered by an arbitrary
number of alternative word or phrase hypotheses.  Using terms of a
Markov Model, a state is allowed to {\em emit a context-free partial
  parse tree}, starting with the represented non-terminal symbol,
yielding part of the sequence of words. This is in contrast to
standard Markov Models. There, states emit atomic symbols.
Note that an edge in the
lattice is represented by a state in the corresponding Markov Model.
Figure \ref{fig:mm-level1}
shows the
part of the Markov Model that represents the best path in the
lattice of figure \ref{fig:cmm-level1}.

The equations of the Viterbi algorithm
are adapted to process a language model
operating on a lattice. Instead of the words, the gaps
between the words are enumerated (see figure \ref{fig:cmm-level1}), and
an edge between two states can span one or more words, such that an
edge is represented by a triple $\langle t,t',q\rangle$, starting at
time $t$, ending at time $t'$ and representing state $q$.

We introduce accumulators $\Delta_{t,t'}(q)$ that collect the maximum
probability of state $q$ covering words from position $t$ to $t'$.  We
use $\delta_{i,j}(q)$ to denote the probability of the deriviation
emitted by state $q$ having a terminal yield that spans positions $i$
to $j$.  These are needed here as part of the accumulators $\Delta$.

\noindent
Initialization:
\begin{equation}
        \Delta_{0,t}(q) = 
                P(q|q_s) \delta_{0,t}(q)
\end{equation}
Recursion:
\begin{equation}
        \Delta_{t,t'}(q) = \max_{\langle
                t'',t,q'\rangle \in \mbox{Lattice}}
                \Delta_{t'',t}(q')P(q|q')
                \delta_{t,t'}(q),
               ~~~
\label{eq:vit-lattice}
\end{equation}
$\mbox{for~}1\leq t < T$.\\
Termination:
\begin{equation}
        \max_{Q\in {\cal Q}^*} P(Q, {\rm Lattice}) = \!\!\!\max_{\langle
                        t,T,q\rangle \in \mbox{Lattice}}\!\!\!
                        \Delta_{t,T}(q) P(q_e|q).
\end{equation}
Additionally, it is necessary to keep track of the elements in the
lattice that maximized each $\Delta_{t,t'}(q)$. When reaching time
$T$, we get the best last element in the lattice
\begin{equation}
        \langle t^m_1,T,q^m_1\rangle = 
        \argmax_{\langle t,T,q\rangle \in \mbox{Lattice}} 
        \Delta_{t,T}(q) P(q_e|q).
\end{equation}
Setting $t_0^m=T$, we collect the arguments $\langle
t'',t,q'\rangle \in \mbox{Lattice}$ that
maximized equation \ref{eq:vit-lattice}
by walking backwards in time:
\begin{equation}
\begin{array}{l}
        \langle t_{i+1}^m, t_i^m, q_{i+1}^m\rangle = \\[1ex]
        \ \argmax\limits_{\langle
                t'',t_i^m,q'\rangle \in \mbox{Lattice}}
                \Delta_{t'',t_i^m}(q') 
  P(q_{i}^m|q') 
                \delta_{t_i^m,t_{i-1}^m}(q_i)
\end{array}
\end{equation}
for $i \geq 1$,
until we reach $t_k^m = 0$. Now, $q^m_1 \ldots q^m_k$ is the best
sequence of phrase hypotheses (read backwards).


\subsection{Passing Ambiguity to the Next Layer}

The process can move on to layer 2 after the first layer is
computed. The results of the first layer are taken as the base and all
context-free rules that apply to the base are retrieved. These again
form a lattice and we can calculate the best path for layer 2.

The Markov Model for layer 1 operates on the output of the Markov
Model for part-of-speech tagging, the model for layer 2 operates on
the output of layer 1, and so on. Hence the name of the processing
model: Cascaded Markov Models.

Very often, it is not sufficient to calculate just the best sequences
of words/tags/phrases. This may result in an error leading to
subsequent errors at higher layers. Therefore, we not only calculate
the best sequence but several top ranked sequences. 
The number of the passed hypotheses depends on a pre-defined threshold
$\theta \geq 1$. We select all hypotheses with probabilities $P \geq
P_{best}/\theta$. These are passed to the next layer together with
their probabilities.


\subsection{Parameter Estimation}

\begin{figure*}
\hrule
\smallskip
\begin{center}
\tabcolsep=2pt\small
\begin{tabular}{c|cccccccccccccc}
Layer & \multicolumn{14}{l}{Sequence}\\
\hline
4 & \multicolumn{14}{c}{\sf S}\\
3 & \multicolumn{7}{c}{\sf NP} & {\sf VAFIN} & \multicolumn{6}{c}{\sf VP}\\
2 & {\sf ART} & {\sf ADJA} & {\sf NN} & \multicolumn{4}{c}{\sf PP} & {\sf VAFIN} & \multicolumn{6}{c}{\sf VP}\\
1 & {\sf ART} & {\sf ADJA} & {\sf NN} & {\sf APPR} & \multicolumn{3}{c}{\sf CNP} &{\sf VAFIN} & \multicolumn{5}{c}{\sf PP} & {\sf VVPP}\\
0 & {\sf ART} & {\sf ADJA} & {\sf NN} & {\sf APPR} & {\sf NN} & {\sf
  KON} & {\sf NN} & {\sf VAFIN} & {\sf APPR} & {\sf ART} & {\sf CARD}
& {\sf ADJA} & {\sf NN} & {\sf VVPP} \\
\end{tabular}

\bigskip

\begin{tabular}{llclc@{\hspace*{1.5cm}}lclcr}
& \multicolumn{6}{l}{Context-free rules and their frequencies} \\
\hline
\hspace*{1em} 
    & {\sf S}  & $\rightarrow$ & {\sf NP VAFIN VP} & (1)
    & {\sf PP} & $\rightarrow$ & {\sf APPR ART CARD ADJA NN} & (1)
    & \hspace*{1em} \\
& {\sf NP} & $\rightarrow$ & {\sf ART ADJA NN PP} & (1) 
    & {\sf ART} & $\rightarrow$ & {\em Ein} & (1) 
    \\
& {\sf PP} & $\rightarrow$ & {\sf APPR CNP} & (1)
    & {\sf ADJA} & $\rightarrow$ & {\em enormer} & (1) 
    \\
& {\sf CNP}& $\rightarrow$ & {\sf NN KON NN} & (1)
    & $\cdots$ & & $\cdots$
    \\
& {\sf VP} & $\rightarrow$ & {\sf PP VVPP} & (1)
    & {\sf VVPP} & $\rightarrow$ & {\em aufgebracht} & (1) 
\end{tabular}

\end{center}
\hrule
\caption{Training material generated from the sentence 
  in figure \ref{fig:examplesent}. The sequences for layers 0
  -- 4 are used to estimate transition probabilities for the
  corresponding Markov Models. The context-free rules are used to
  estimate the SCFG, which determines the output probabilities of the
  Markov Models.}
\label{fig:training}
\end{figure*}

Transitional parameters for Cascaded Markov Models are estimated
separately for each layer. Output parameters are the same for all
layers, they are taken from the stochastic context-free grammar that is
read off the treebank.

Training on annotated data is straight forward.
First, we number the layers, starting with 0 for the part-of-speech
layer. Subsequently, information for the different layers is collected.

Each sentence in the corpus represents one training sequence for each
layer. This sequence consists of the tags or phrases at that layer.  If
a span is not covered by a phrase at a particular layer, we take the
elements of the highest layer below the actual layer. Figure
\ref{fig:training} shows the training sequences for layers 0 -- 4
generated from the sentence in figure \ref{fig:examplesent}. Each
sentence gives rise to one training sequence for each layer.
Contextual parameter estimation is done in analogy to models for
part-of-speech tagging, and the same smoothing techniques can be
applied. We use a linear interpolation of uni-, bi-, and trigram
models.

A stochastic context-free grammar is read off the corpus. The rules
derived from the annotated sentence in figure \ref{fig:examplesent} are
also shown in figure \ref{fig:training}. The grammar is used to
estimate output parameters for all Markov Models, i.e., they are the
same for all layers. We could estimate probabilities for rules
separately for each layer, but this would worsen the sparse data
problem.


\section{Experiments}
\label{sec:experiment}

\begin{figure*}[t]
\hrule
\bigskip
\psfig{file=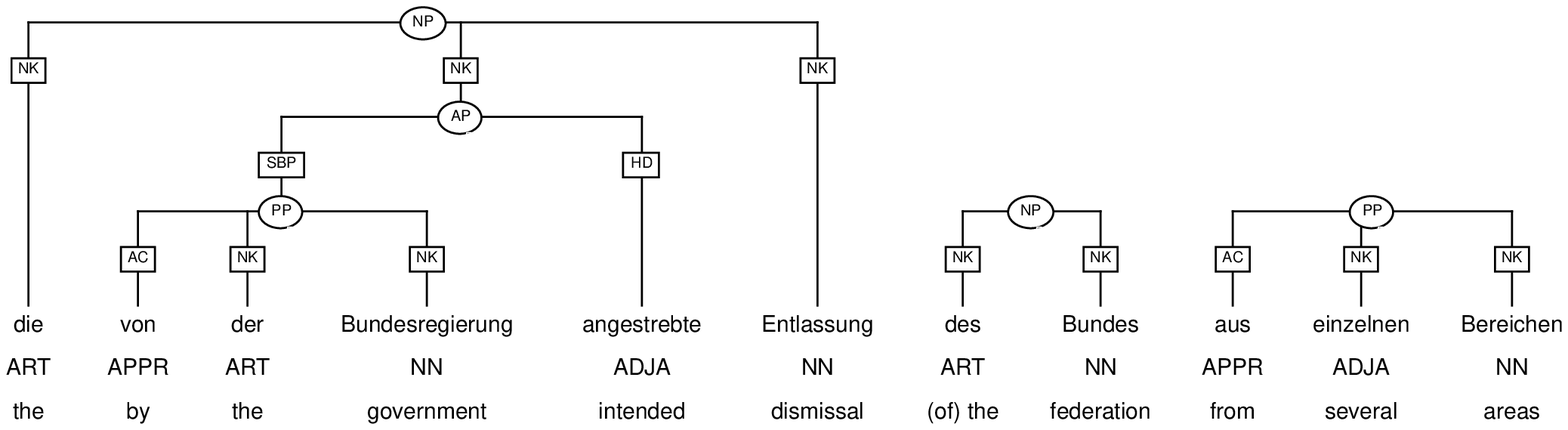,width=\textwidth}
\vspace*{-2ex}
\centerline{\em `the dismissal of the federation from several
  areas that was intended by the government'}
\vspace{.3ex}
\hrule
\caption{Complex German {\sf NP} and chunker output (postnominal
  genitive and {\sf PP} are not attached).}
\label{fig:compnp}
\bigskip
\hrule
\bigskip
\begin{center}
{\bf NEGRA Corpus: Chunking Results}
\bigskip
\hspace*{0pt}
\beginpicture
\setcoordinatesystem units <10mm,0.76mm>
\setplotarea x from 1 to 9, y from 60 to 100
\axis bottom ticks numbered from 1 to 9 by 1 /
\axis left ticks numbered from 60 to 100 by 10 /
\linethickness=.4pt
\putbar breadth <0pt> from 1 70 to 9 70
\putbar breadth <0pt> from 1 80 to 9 80
\putbar breadth <0pt> from 1 90 to 9 90
\putbar breadth <0pt> from 1 100 to 9 100
\put {\rotate[l]{Recall/Precision}} [r] <-9mm,0mm> at 1 80
\put {\# Layers} [lt] <5mm,-2.7mm> at 9 60
\put {\small 96.2} [t] <0mm,-8.5mm> at 1 60
\put {\small 96.3} [t] <0mm,-8.5mm> at 2 60 
\put {\small 96.4} [t] <0mm,-8.5mm> at 3 60 
\put {\small 96.4} [t] <0mm,-8.5mm> at 4 60 
\put {\small 96.5} [t] <0mm,-8.5mm> at 5 60 
\put {\small 96.5} [t] <0mm,-8.5mm> at 6 60 
\put {\small 96.5} [t] <0mm,-8.5mm> at 7 60 
\put {\small 96.5} [t] <0mm,-8.5mm> at 8 60 
\put {\small 96.5} [t] <0mm,-8.5mm> at 9 60 
\put {\small \% POS accuracy} [lt] <5mm,-8mm> at 9 60
\put {\tabcolsep=0pt\begin{tabular}{llcr}
    \setgray{0.5}\rule[.5ex]{10mm}{.8pt}\unsetgray
                             & \multicolumn{3}{l}{Topline Recall} \\
                             & \small min &\small=&\small 72.6\% \\
                             & \small max &\small=&\small 100.0\% \\
    \hbox to 0pt{\hspace*{4mm}$\bullet$\hss}%
    \rule[.5ex]{10mm}{.8pt}\ \ \ & \multicolumn{3}{l}{Recall} \\
                             & \small min &\small=&\small 54.0\% \\
                             & \small max &\small=&\small 84.8\% \\
    \hbox to 0pt{\hspace*{4mm}$\circ$\hss}%
    \rule[.5ex]{10mm}{.8pt} & \multicolumn{3}{l}{Precision} \\
                             & \small min &\small=&\small 88.3\% \\
                             & \small max &\small=&\small 91.4\% \\
    \end{tabular}} [lb] <5mm,0mm> at 9 60
\setlinear


\setplotsymbol({\rule{0.8pt}{0.8pt}})

\setgray{0.5}
\plot
        1       72.55 
        2       92.54 
        3       98.09 
        4       99.58 
        5       99.88 
        6       99.95 
        7       99.98 
        8      100.00 
        9      100.00 
/
\unsetgray%


\setplotsymbol({\rule{.8pt}{.8pt}})
\plot
        1.194   58
        2       74.73 
        3       80.79 
        4       82.84 
        5       83.97 
        6       84.40 
        7       84.71 
        8       84.74 
        9       84.75 
/
\multiput {$\bullet$} at
        2       74.73 
        3       80.79 
        4       82.84 
        5       83.97 
        6       84.40 
        7       84.71 
        8       84.74 
        9       84.75 
/

\setplotsymbol({\rule{.8pt}{.8pt}})
\plot
        1       91.35 
        2       90.40 
        3       89.51 
        4       89.03 
        5       88.71 
        6       88.46 
        7       88.34 
        8       88.29 
        9       88.27 
/
\multiput {$\circ$} at
        1       91.35 
        2       90.40 
        3       89.51 
        4       89.03 
        5       88.71 
        6       88.46 
        7       88.34 
        8       88.29 
        9       88.27 
/

\endpicture
\end{center}
\vspace{-5mm}
\hrule
\caption{{\sf NP}/{\sf PP} chunking results for the NEGRA Corpus.
  The diagram shows recall and precision depending on the number of
  layers that are used for parsing. Layer 0 is used for part-of-speech
  tagging, for which tagging accuracies are given at the bottom
  line. Topline recall is the maximum recall possible for that number
  of layers.}
\label{fig:negra-parse}
\end{figure*}

This section reports on results of experiments with Cascaded
Markov Models. 
We evaluate chunking precision and recall, i.e., the recognition
of kernel {\sf NP}s and {\sf PP}s. These exclude prenominal adverbs
and postnominal {\sf PP}s and relative clauses, but include all other
prenominal modifiers, which can be fairly complex adjective phrases in
German. Figure \ref{fig:compnp} shows an example of a complex
{\sf NP} and the output of the parsing process.

For our experiments, we use the NEGRA corpus \cite{Skut:ea:97a}. It
consists of German newspaper texts (Frankfurter Rundschau) that are
annotated with predicate-argument structures. We
extracted all structures for {\sf NP}s, {\sf PP}s, {\sf AP}s,
{\sf AVP}s (i.e., we mainly excluded sentences, {\sf
  VP}s and coordinations). The version of the corpus used contains
17,000 sentences (300,000 tokens).

The corpus was divided into training part (90\%) and test part (10\%).
Experiments were repeated 10 times, results were averaged.
Cross-evaluation was done in order to obtain more reliable
performance estimates than by just one test run.  \mbox{Input} of the process
is a sequence of words (\mbox{divided} into sentences), output are
part-of-speech tags and structures like the one indicated in figure
\ref{fig:compnp}.

Figure \ref{fig:negra-parse} presents results of the chunking task using
Cascaded Markov Models for different numbers of layers.\footnote{The
  figure indicates unlabeled recall and precision. Differences to
  labeled recall/precision are small, since the number of different
  non-terminal categories is very restricted.}  Percentages are
slightly below those presented by \cite{Skut:Brants:98a}. But they
started with correctly tagged data, so our task is harder since it includes the
process of part-of-speech tagging.

Recall increases with the number of layers. It ranges from 54.0\% for 1
layer to 84.8\% for 9 layers. This could be expected,
because the number of layers determines the number of phrases that can
be parsed by the model. The additional line for ``topline recall''
indicates the percentage of phrases that can be parsed by Cascaded
Markov Models with the given number of layers. All nodes that belong to
higher layers cannot be recognized.

Precision slightly decreases with the number of layers. It ranges from
91.4\% for 1 layer to 88.3\% for 9 layers.

The $F$-score is a weighted combination of recall $R$ and precision
$P$ and defined as follows: 
\begin{equation}
  F = \frac{(\beta^2 + 1) PR}{\beta^2P + R}
\end{equation}
$\beta$ is a parameter encoding the importance of recall and
precision. Using an equal weight for both ($\beta =
1$), the maximum $F$-score is reached for 7 layers ($F = $86.5\%).

The part-of-speech tagging accuracy slightly increases with
the number of Markov Model layers (bottom line in figure
\ref{fig:negra-parse}).
This can be explained by top-down decisions of Cascaded Markov Models.
A model at a higher layer can select a tag with a lower probability if
this increases the probability at that layer. Thereby some errors made
at lower layers can be corrected. This leads to the increase of up to 
0.3\% in accuracy.

Results for chunking Penn Treebank data were previously presented by
several authors
\cite{Ramshaw:Marcus:95,Argamon:ea:98,Veenstra:98,Cardie:Pierce:98}.
These are not directly comparable to our results, because they
processed a different language and generated only one layer of
structure (the chunk boundaries), while our algorithm also generates
the internal structure of chunks. But generally, Cascaded Markov Models
can be reduced to generating just one layer and can be trained on Penn
Treebank data.


\section{Conclusion and Future Work}
\label{sec:conclusion}

We have presented a new parsing model for shallow processing. The model
parses by representing each layer of the resulting structure as a
separate Markov Model. States represent categories of words and
phrases, outputs consist of partial parse trees. Starting with the
layer for part-of-speech tags, the output of lower layers is passed as
input to higher layers. This type of model is restricted to a fixed
maximum number of layers in the parsed structure, since the number of
Markov Models is determined before parsing. While the effects of these
restrictions on the parsing of sentences and {\sf VP}s are still to be
investigated, we obtain excellent results for the chunking task, i.e.,
the recognition of kernel {\sf NP}s and {\sf PP}s.

It will be interesting to see in future work if Cascaded Markov Models
can be extended to parsing sentences and {\sf VP}s. The average number
of layers per sentence in the NEGRA corpus is only 5; 99.9\% of all
sentences have 10 or less layers, thus a very limited number of Markov
Models would be sufficient.

Cascaded Markov Models add left-to-right context-information to
context-free parsing. This {\em contextualization\/} is orthogonal to
another important trend in language processing: {\em lexicalization}.
We expect that the combination of these techniques results in improved
models.

We presented the generation of parameters from annotated corpora and
used linear interpolation for smoothing. While we do not expect
improvements by re-estimation on raw data, other smoothing methods may
result in better accuracies, e.g. the maximum entropy framework.  Yet,
the high complexity of maximum entropy parameter estimation requires
careful pre-selection of relevant linguistic features.

The presented Markov Models act as filters. The probability of
the resulting structure is determined only based on a stochastic
context-free grammar. While building the structure bottom up, parses
that are unlikely according to the Markov Models are pruned. We think
that a combined probability measure would improve the model. For this,
a mathematically motivated combination needs to be determined.



\section*{Acknowledgements}

I would like to thank Hans Uszkoreit, Yves \mbox{Schabes}, Wojciech Skut, and
Matthew Crocker for fruitful discussions and valuable comments on the
work presented here.  And I am grateful to Sabine Kramp for
proof-reading this paper.

This research was funded by the Deutsche Forschungsgemeinschaft in the
Sonderforschungsbereich 378, Project C3 NEGRA.


\makeatletter
\write\@auxout{\string\gdef\string\lastpage{\thepage}}
\makeatother

\end{document}